\documentclass[conference]{IEEEtran}
\IEEEoverridecommandlockouts
\usepackage{cite}
\usepackage{amsmath,amssymb,amsfonts}
\usepackage{algorithmic}
\usepackage{graphicx}
\usepackage{textcomp}
\usepackage{xcolor}
\usepackage{amsmath}
\usepackage{mathtools}
\usepackage{graphicx}
\usepackage{url}
\usepackage{amsfonts}
\usepackage{amssymb}
\usepackage{color}
\usepackage{xfrac}
\usepackage{siunitx}
\usepackage{float}
\usepackage{listings}
\usepackage{algorithm}
\usepackage{bm}
\usepackage{changes}
\usepackage{booktabs}
\def\BibTeX{{\rm B\kern-.05em{\sc i\kern-.025em b}\kern-.08em
    T\kern-.1667em\lower.7ex\hbox{E}\kern-.125emX}}
    
\usepackage{tikz}
\usetikzlibrary{shapes, arrows}
\tikzstyle{block} = [draw, fill=blue!20, rectangle, 
    minimum height=3em, minimum width=3em, node distance=2cm]
\tikzstyle{block2} = [draw, fill=blue!20, rectangle, 
    minimum height=3em, minimum width=3em, node distance=1.5cm]
\tikzstyle{sum} = [draw, fill=blue!20, circle, node distance=1.5cm]
\tikzstyle{input} = [coordinate]
\tikzstyle{output} = [coordinate]
\tikzstyle{pinstyle} = [pin edge={to-,thin,black}]
\tikzstyle{skip} = [coordinate, node distance=3cm]

\begin{document}

\title{OpinionRank: Extracting Ground Truth Labels from Unreliable Expert Opinions with Graph-Based Spectral Ranking
}

\author{\IEEEauthorblockN{Glenn Dawson}
\IEEEauthorblockA{\textit{Rowan University} \\
Glassboro, New Jersey, United States \\
dawsong@rowan.edu}
\and
\IEEEauthorblockN{Robi Polikar}
\IEEEauthorblockA{\textit{Rowan University}\\
Glassboro, New Jersey, United States\\
polikar@rowan.edu}
}

\maketitle

\begin{abstract}
     As larger and more comprehensive datasets become standard in contemporary machine learning, it becomes increasingly more difficult to obtain reliable, trustworthy label information with which to train sophisticated models. To address this problem, crowdsourcing has emerged as a popular, inexpensive, and efficient data mining solution for performing distributed label collection. However, crowdsourced annotations are inherently untrustworthy, as the labels are provided by anonymous volunteers who may have varying, unreliable expertise. Worse yet, some participants on commonly used platforms such as Amazon Mechanical Turk may be adversarial, and provide intentionally incorrect label information without the end user's knowledge. We discuss three conventional models of the label generation process, describing their parameterizations and the model-based approaches used to solve them. We then propose OpinionRank, a model-free, interpretable, graph-based spectral algorithm for integrating crowdsourced annotations into reliable labels for performing supervised or semi-supervised learning. Our experiments show that OpinionRank performs favorably when compared against more highly parameterized algorithms. We also show that OpinionRank is scalable to very large datasets and numbers of label sources, and requires considerably fewer computational resources than previous approaches.
\end{abstract}


\section{Introduction}
Supervised machine learning is built on the fundamental assumption that the labels provided by the training and testing datasets are accurate. Regardless of the application, from data mining to computer vision to natural language processing, trust in the ground truth provided by the datasets is paramount. However, increasingly many of these applications use increasingly large datasets (sometimes on the order of millions or even billions of examples); such datasets are simply far too large to annotate manually. Crowdsourcing has therefore emerged as an appealing, inexpensive tool for distributed collection of label information for large-scale datasets; the widely used ImageNet dataset, for example, is annotated using crowdsourcing from Amazon Mechanical Turk \cite{imagenet}. Crowdsourcing is also widely used to annotate datasets in natural language processing \cite{snow2008cheap, wang2013} and data mining \cite{guo2019, sheng2008, xintong2014}.

However, crowdsourcing suffers from the problem of inexpert---and therefore unreliable---labeling. The very property of widespread contribution that gives crowdsourcing its power also results in the significant drawback of relying on the opinions of ``experts'' (sometimes referred to as labelers, annotators, workers, or label sources) who may have little or no domain knowledge. Furthermore, there may be differences of opinion between labelers with contrasting expertise: one expert may provide a label that could be considered correct from one perspective that disagrees with a label that could be considered correct from a different perspective. For example, recent works that have analyzed the ImageNet dataset found that using the crowdsourced labels as a gold standard may be flawed \cite{imagenet-done, imagenet-relabeling}. Perhaps more importantly, there may also be labels provided by adversarial labelers, who intentionally provide incorrect labels with the sole purpose of confounding the learner.  Finally, we cannot discount the possibility of corruption between the labels provided by a source and the labels observed by a learner, examples of which may include recording errors, data transfer errors, modeling errors, or simply missing labels. 

Against this background of accumulating uncertainties, and the fact that any particular label source must therefore be treated with some skepticism, our goal is to determine which sources---and to what extent---should be considered in assigning labels, particularly when there are conflicting opinions. In the absence of unassailable ground truth labels, it is necessary to solve the unsupervised problem of integrating unreliable expert opinions into reliable ground truth labels.

\section{Related Work}
Perhaps the earliest work on characterizing the collective decision of a set of inexpert opinions is the Condorcet jury theorem \cite{condorcet}, which states that for a group of independent voters with a homogeneous probability of correctness $p$, the probability of their majority vote being correct on a binary decision increases with the size of the group if $p > 0.5$. Kazmann showed that the assumption of homogeneous voters can be relaxed by assuming that the heterogeneous voter correctness probabilities follow a symmetrical distribution about a mean $\overline{p}$ \cite{democratic}. Grofman demonstrated that the group's collective accuracy can increase even if the added members are less competent than the group's previous average \cite{condorcet-unreliable}. Owen, Grofman, and Feld removed the distribution restrictions, generalizing the theorem to depend only on $\overline{p}>0.5$ \cite{condorcet-extended}. List and Goodin extended this result to problems with more than two classes, showing that for a $k$-class problem, the average voter reliability needs only to exceed $1 / k$ for the majority vote decision to be correct \cite{condorcet-multiclass}.

Beyond simple majority vote analysis, and the associated large body of ensemble-based approaches, much research has gone into investigating crowdsourcing algorithms. Dawid and Skene proposed a model based on the well-known expectation-maximization (EM) algorithm, attempting to estimate each annotator's respective expertise as a confusion matrix \cite{dawid1979maximum}. From their application, a rich body of work has sprouted, with many improvements, alterations, and theoretical bounds on the performance of generative models based on the EM approach \cite{snow2008cheap, venus, whitehill, ensemble-linear, zhou2014aggregating, gao2013minimax, rodrigues2013, yan2014learning, zhang2018}. Aside from EM-based approaches, Ghosh et al. \cite{ghosh} and Dalvi et al. \cite{dalvi} proposed sparse matrix algorithms based on singular value decomposition. Other approaches have investigated Bayesian inference \cite{chen, liu, raykar, welinder}, as well as spectral methods \cite{zou, zhang}. An intriguing line of investigation by Goldberger examined the problem of inexact labels \cite{goldberger}. More recently, the success of deep neural networks has prompted work on deep generative models \cite{yin2017, atarashi2018semi, shi2020}. Regardless of the specific approach, nearly all work in this area attempts to model either the parameters of annotators' reliabilities, or the confusion matrices associated with each annotator. 

Ranking algorithms are also commonly used in applications such as information retrieval, bioinformatics, and recommender systems. Most well-known is the PageRank algorithm, which utilizes the power iteration method to rank web pages based on how frequently other web pages link to them \cite{pagerank}. Similar graph-based methods are used in natural language processing for extractive document summarization by ranking sentence importance \cite{mihalcea2004textrank, erkan2004lexrank}. Other approaches include gradient-based techniques \cite{burges2005learning}, multi-criteria max-margin algorithms \cite{chakrabarti2008structured}, and support vector methods \cite{yue2007support}. 

\section{Parameterized Models for Label Generation}
\label{settings}

In this section, we review and discuss three models of learning from unreliable sources. The first two settings are well-established and widely cited, and can be considered to be canonical models. While more recent work has proposed some additions to these models, the fundamental idea of highly parameterized probabilistic dynamics is consistent across the literature. The third model adds a new layer of complexity to the label generating process that cannot be easily incorporated into the previous models, and so we give it special consideration. For consistency, we use the original authors' terminologies for label assigners (such as ``labelers'', ``annotators'', or ``experts''); each of these terms is understood to refer to the same concept.

\subsection{Joint Modeling of Source Reliability and Instance Difficulty}
\label{whitehill-setting}
We first consider the setting proposed by Whitehill et al. \cite{whitehill}, who model the labeling process as not only a function of the \textit{expertise} of the labeler, parameterized by $\alpha \in (-\infty, \infty)$, but also as the \textit{difficulty} of labeling a data instance, parameterized by $1 / \beta \in [0, \infty)$. When $1/\beta = \infty$, an instance is deemed very difficult to label for even the most expert labeler, whereas $1/\beta = 0$ represents a trivial instance, so obvious that anyone would label it correctly irrespective of expertise. The range of values for $\alpha$ describes expertise from ``perfectly wrong'' (when $\alpha = -\infty$) to ``perfectly accurate'' (when $\alpha = \infty$), with $\alpha = 0$ representing random guessing. The probability of the label $L_{ij}$---assigned by labeler $j$ on instance $i$---being the correct label $Z_i \in [0, 1]$ is then modeled as \begin{displaymath}
    \Pr(L_{ij} = Z_i|\alpha_j, \beta_i) = \frac{1}{1 + e^{-\alpha_j\beta_i}}.
\end{displaymath}

The model allows the log odds for the label being correct to be expressed as \begin{displaymath}
    \log\frac{\Pr(L_{ij} = Z_i)}{1 - \Pr(L_{ij} = Z_i)} = \alpha_j\beta_i.
\end{displaymath}

From this formulation, the authors develop an EM-based algorithm called GLAD (\textbf{G}enerative model of \textbf{L}abels, \textbf{A}bilities, and \textbf{D}ifficulties), which---under the assumptions of the labeling model---is able to recover $Z$, $\alpha$, and $\beta$ for all data and labelers.

To our knowledge, Whitehill et al. were the first to extend the work of Dawid and Skene (who assumed that labels were generated only by parameters over the labelers' expertise \cite{dawid1979maximum}) such that the generative model also included parameters for the difficulty of assigning the correct label. While adding a small amount of complexity, Whitehill et al. sought to keep their model as simple as possible, assigning only a single hidden parameter for each instance and labeler.

\subsection{Multidimensional Parameterization of Label Generation}
\label{welinder-setting}
    
Welinder et al. considered the annotation process to be a high-dimensional system, with both instance difficulty and annotator ability being governed by many parameters \cite{welinder}. For their formulation, the authors suggest that the dynamics can be described as a Gaussian mixture model. For each instance $i$, with ground truth label $z_i \in [0, 1]$, the presentation $x_i$ of instance to annotator $j$ is modeled by 
\begin{gather*}
    x_i\sim\mathcal{N}(\mu_z, \theta_z^2)\\
    \mu_z = \begin{cases}
                -1 &\mathrm{if}\ z_i = 0\\
                1 &\mathrm{if}\ z_i = 1
            \end{cases},
\end{gather*}
\noindent where $\theta_z$ is a parameter describing the variability in the difficulty of correctly labeling instance $i$. Annotator $j$ sees a version of instance $i$ modeled as $y_{ij} = x_i + n_{ij}$, where $n_{ij}$ is the annotator- and instance-specific ``noise'', such as differences in annotator attention, acuity, direction of gaze, etc. The noise statistics vary from annotator to annotator, and are modeled as a parameter $\sigma_j$. The authors assume Gaussian noise, i.e. $y_{ij} \sim\mathcal{N}(x_i, \sigma_j^2)$. The annotator-assigned label $L_{ij}$ is then chosen deterministically as $L_{ij} = \mathbb{I}(\langle\hat{w}_j, y_{ij}\rangle \geq \hat{\tau}_j)$, where $\mathbb{I}(\cdot)$ is the indicator function and $\hat{w}_j$ is a weighting vector that encodes each annotator's expertise. The authors draw the decision threshold according to $\tau_j$ following a zero-mean Gaussian, and sample the noise parameter $\sigma_j$ from a gamma distribution. 

The authors then apply Bayesian maximum a posteriori (MAP) estimation to maximize the posterior on the parameters. They solve this optimization using gradient ascent by alternating between fixing $x$ and optimizing over ($w, \tau$), and fixing ($w, \tau$) and optimizing $x$, assuming Gaussian priors on $w_j$ and $\tau_j$, respectively.





The generative model proposed by Welinder et al. is highly parameterized and considerably more complex than the model suggested by Whitehill et al. Later models build on this idea, such as that of Atarashi et al., adding even more parameterization in the form of latent variables  \cite{atarashi2018semi}. Latent variable models are intractable for EM algorithms; more recent authors have turned to deep neural networks for approximating these more complex environments \cite{yin2017, atarashi2018semi, shi2020}. Each of these models, however, shares lineage with the work of Welinder et al., and the general \textit{class} of multidimensional label generation dynamics is well-represented by their work.

\subsection{``Soft'' Observations of Expert Opinions}
\label{goldberger-setting}
    
Goldberger introduced the notion of soft opinions, where categorical label assignment of expert $j$ on data instance $i$ (with true label $z_i$) is not a one-hot encoding, but rather a probability distribution \cite{goldberger}. He simplified the initial annotation process compared to the previous models, assuming only that the expert's initial opinion is assigned following
\begin{displaymath}
    \Pr(y_{ij}|z_i = a; p_{j}) = \begin{rcases}
        \begin{dcases}
            \ \ \ p_j, &\text{ if } y_{ij} = a\\
            \frac{1 - p_j}{|A| - 1}, &\text{ if } y_{ij} \neq a
        \end{dcases}
    \end{rcases},\ \ \forall a \in A,
\end{displaymath}
where $p_j$ is the expert's probability of providing the correct label, and $A$ is the set of possible labels. Goldberger assumed that each source has an identical reliability across all classes, and that an incorrect label is assigned following a uniform distribution across the incorrect classes. 

Goldberger's most notable contribution is that he extends the uncertainty in observed labels beyond that of the expert's label generation model. He assumes that the \textit{observed} version of the expert opinion, $q_{ij}$, is not an indicator, but rather a probability distribution over all visible labels:
\begin{displaymath}
    q_{ij}(b) = \Pr(y_{ij} = b),\ \ \forall b \in A.
\end{displaymath}

In this way, Goldberger accounts for a layer of obfuscation between the labels as provided by the experts and the labels as seen by the observer. This consideration is intriguing, and adds an important contribution that is missing from the previous models, extending the assumption of unreliability from simply the \textit{generation} of labels to the \textit{observation} of the labels. Goldberger handles this obfuscation by developing an extended EM algorithm.

\section{OpinionRank: Spectral Ranking With Corroboration Graphs}
Previous approaches using expectation-maximization have been shown to be highly effective for estimating label-generating dynamics, under the assumption that these dynamics follow the specific parameterizations of the model. However, EM approaches depend highly upon the correct parameterization of the system, and can fail in alternative environments. Similarly, deep generative models have achieved remarkable results in semi-supervised settings, but they too rely on correct parameterization, in addition to their lack of interpretability. Finally, both EM approaches and deep neural networks demand substantial computational requirements to converge to their estimates of the system dynamics. 

To address these drawbacks, we propose OpinionRank, a spectral algorithm for expert ranking and weighted voting. Instead of attempting to estimate the \textit{true} reliability parameters of each expert in the ensemble, we propose to estimate the \textit{relative} reliability of each expert. Furthermore, we do so using a nonparameterized approach: given only the observed labels (of unknown reliability) provided by each expert, we compute our estimation of the experts' relative expertise by comparing the \textit{frequency of agreement} between each pair of experts. An agreement between two experts can be interpreted as a soft ``recommendation'' between them: given that they have provided the same label for the same instance, it is reasonable to expect that one expert would recommend the other at least some of the time. Under the Condorcet criterion that the average expertise of the ensemble of experts exceeds random guessing, this system of mutual recommendations builds a network of trust. We formulate the ensemble as a fully connected graph, with each expert functioning as a node. The edges of each node correspond to the number of times that each expert $i$ agrees with (recommends) each other expert $j$ (an expert always recommends itself). 

We consider the probabilistic interpretation of the frequency of expert $i$ recommending any other expert $j$. The edges leading outward from any expert $i$ can be transformed into a probability distribution; we interpret the graph of interexpert agreements as a Markov chain. The edge probabilities of recommendation form a dense transition matrix, which we call a \textit{corroboration} matrix. We guarantee that the corroboration matrix is ergodic and irreducible by selecting the softmax function to form the edge distributions. For stability, we scale the entire corroboration matrix by dividing by the total number of examples before applying the softmax operation.

The Perron-Frobenius theorem guarantees that the corroboration matrix---being real, square, and positive---will have a unique, positive eigenvector \cite{frobenius1912matrizen}. This dominant eigenvector represents the steady-state probabilities of the corroboration matrix, with its elements describing the long-run probabilities of choosing the opinion of any given expert after stochastically asking each expert to recommend other experts. We compute the dominant eigenvector $\bm{v}$ using the well-known power iteration method, raising the transformed corroboration matrix $\mathcal{C}$ to an arbitrarily high matrix power $P$ and multiplying by an elementary vector $\text{\bf{e}}$:

\begin{equation*}
    \label{eq:power-iteration}
    \bm{v} = (\mathcal{C}^P)^\intercal \text{\bf{e}}.
\end{equation*}

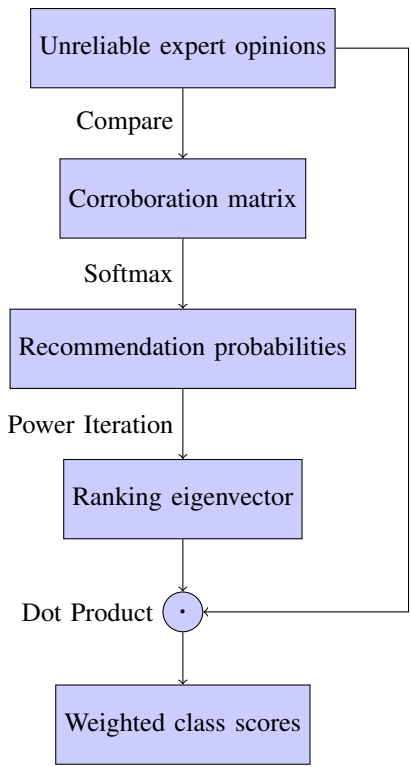
\begin{figure}
    \centering
    \begin{tikzpicture}[auto, node distance=2cm]
        \node [block, name=opinions] (opinions) {Unreliable expert opinions};
        \node [block, below of=opinions] (corroboration) {Corroboration matrix};
        \node [block, below of=corroboration] (softmax) {Recommendation probabilities};
        \node [block, below of=softmax] (eigen) {Ranking eigenvector};
        \node [sum, below of=eigen] (dot) {$\bm{\cdot}$};
        \node [left] at (dot.west) {Dot Product};
        \node [block2, below of=dot] (labels) {Weighted class scores};
        \coordinate [skip, right of=opinions] (tmp) {};
        \coordinate [skip, right of=dot] (tmp2) {};
        
        \draw [->] (opinions) -- node[left] {Compare} (corroboration);
        \draw [->] (corroboration) -- node[left] {Softmax} (softmax);
        \draw [->] (softmax) -- node[left] {Power Iteration} (eigen);
        \draw [->] (eigen) -- node {} (dot);
        \draw [->] (opinions) |- (tmp) -- (tmp2) -- node {} (dot);
        \draw [->] (dot) -- node{} (labels);
    \end{tikzpicture}
    \caption{Block diagram of OpinionRank.}
    \label{opinionrank-block}
\end{figure}

The use of the Perron-Frobenius eigenvector as a ranking tool is most often associated with the PageRank algorithm \cite{pagerank}, though its use in this application goes back even further \cite{seeley1949net, Keener1993ThePT}. Recent theoretical work has shown that under mild assumptions about the underlying properties of the objects being ranked, the spectral method of eigenvector ranking is equally optimal as maximum likelihood estimation approaches \cite{chen2019spectral}. Here, we interpret the probabilities of the eigenvector as a scheme for weighting the votes of each expert. For each instance, we take the dot product between the binary vector of expert opinions on class membership and the relative reliability vector to produce a scalar value $w \in [0, 1]$ representing the weighted ensemble opinion on the class membership of the example. Optionally, we can choose to treat the eigenvector as a strict ranking, and retain only the top-$N$ experts. We summarize OpinionRank in Algorithm \ref{alg:opinionrank}; a visual diagram outlining the algorithm's flow is shown in Figure \ref{opinionrank-block}.

\begin{algorithm}
    \caption{OpinionRank: A Model-Free, Graph-Based Spectral Method for Extracting Labels from Unreliable Expert Opinions}
    \label{alg:opinionrank}
    \begin{algorithmic}[1]
        \renewcommand{\algorithmicrequire}{\textbf{Input:}}
        \renewcommand{\algorithmicensure}{\textbf{Output:}}
        \REQUIRE $\mathcal{Y}$, a set of $k$-class membership opinions on $n$ total examples from $s$ sources.
        \REQUIRE $P$, the number of matrix power iterations
        \REQUIRE $N \in [1, s]$, the number of sources to retain from the dominant eigenvector
        \ENSURE $\mathcal{W}$, a $k \times n$ matrix of weighted class membership scores.
        \STATE Initialize $k \times n$ matrix of class scores, $\mathcal{W}$
        \FOR {each class $\ell=1$ to $k$}
            \STATE Obtain $s \times n$ matrix of class-$\ell$ membership opinions, 
            \\ $\mathcal{K} \leftarrow \text{checkEqual}(\mathcal{Y}, \ell)$
            \STATE Initialize $s \times s$ corroboration matrix $\mathcal{C}$ 
            \FOR {each source $i=1$ to $s$}
                \FOR {each source $j=1$ to $s$}
                    \STATE $\mathcal{C}_{ij} \leftarrow \sum \text{checkEqual}(\mathcal{K}_i, \mathcal{K}_j)$
                \ENDFOR
                \STATE $\mathcal{C}_i \leftarrow \text{softmax}(\mathcal{C}_i / n)$
            \ENDFOR
            \STATE Obtain dominant eigenvector, $\bm{v} \leftarrow (\mathcal{C}^P)^\intercal\text{\textbf{e}}$
            \STATE Obtain top-$N$ eigenvector indices, $\bm{I} \leftarrow \text{argsort}(\bm{v}, N)$
            \FOR {all $k \in \mathcal{K}$}
                \STATE Replace missing values, $k \leftarrow 0.5$ if $k$ is missing
            \ENDFOR
            \STATE $\mathcal{W}_\ell \leftarrow \mathcal{K}_I^\intercal \bm{v}_I$
        \ENDFOR
        \RETURN $\mathcal{W}$
    \end{algorithmic}
\end{algorithm}

The OpinionRank algorithm is highly flexible, and is easily adaptable to any labeling paradigm. In the case of binary categorical labeling problems, OpinionRank can be applied directly. For multinomial and multilabel problems, the expert-provided class labels can be transformed into binary encodings (one-hot labels for the multinomial case), with OpinionRank being applied across each class. In these scenarios, OpinionRank estimates the class-conditional reliability ranking of each expert, on the observation that some experts may have more or less expertise with respect to some classes compared to others. 

For binary problems, label predictions are obtained by thresholding the weighted class scores at 0.5. In the multilabel case, the same rule can be applied to each class independently to obtain binary label vectors for each instance. Multinomial decisions are made by choosing the class corresponding to the argmax of the class membership scores. 

OpinionRank can also easily handle missing values, such as the case where some experts did not label every instance. We consider an expert that does not provide a label for a given instance to not be in agreement with any other expert, including itself, for that instance. A missing value therefore reduces the overall reliability of an expert, as there is less evidence that the expert is trustworthy. When performing inference, missing values are replaced with 0.5, so that they do not bias the final decision of the algorithm. 




\section{Experimental Design and Results}
\label{experimental-design}
In order to objectively evaluate and compare our proposed algorithm, we reproduce, as faithfully as possible, experiments from each of the settings described in Section \ref{settings}, and test the OpinionRank algorithm under the hand-crafted conditions for the models of the original authors. We also perform a wall clock runtime analysis of OpinionRank to demonstrate the algorithm's speed and computational efficiency.

\subsection{Whose Vote Should Count More: Optimal Integration of Labels from Labelers of Unknown Expertise}

We implemented three experiments under the same conditions described in Whitehill et al. \cite{whitehill}. These experiments evaluate the OpinionRank algorithm's performance under the conditions of the authors' labeling model (as described in Section \ref{whitehill-setting}), its ability to handle ``difficult'' images, and its stability under varying starting conditions.

\subsubsection{Labeling Model}
\label{whitehill-ex-1}
\begin{figure}
    \centering
    \includegraphics[width=\linewidth]{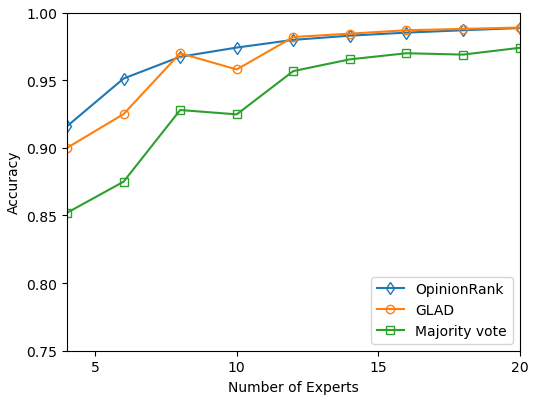}
    \caption{The accuracy of OpinionRank versus the GLAD model and simple majority voting for inferring underlying class labels on the binary labeling experiment (Section \ref{whitehill-ex-1}). Baseline results reproduced from Figure 2 in Whitehill et al. \cite{whitehill}.}
    \label{whitehill2009-1}
\end{figure}

The first experiment simulates the labeler accuracy as $\alpha_j \sim \mathcal{N}(1, 1)$, and the inverse-difficulty of labeling an image as $\beta_i \sim \text{Lognormal}(1, 1)$. The observed label of an instance $i$ provided by labeler $j$ is sampled according to the probability $\Pr(L_{ij} = Z_i|\alpha_j, \beta_i) = \frac{1}{1 + e^{-\alpha_j\beta_i}}$. The algorithms are evaluated by the proportion of accurate class labels, with the amount of total data set to $n = 200$. Whitehill et al. reported the average of 40 experiments; we report the mean of 50,000 experiments (Figure \ref{whitehill2009-1}).

Both OpinionRank and GLAD considerably outperform majority voting, and converge to $>99\%$ accuracy as the number of labelers increases. Because the parameters of the experiment ensure that the average reliability of the pool of labelers is greater than 0.5, these results are expected (from the generalized Condorcet jury theorem). Notably, OpinionRank outperforms GLAD at lower numbers of labelers, suggesting that eigenvector-based reliability ranking is robust even for small pools of labelers.




\subsubsection{Modeling Image Difficulty}
\label{whitehill-ex-2}
The second experiment considers a pool of 50 labelers, each labeling the same set of $n=1000$ instances, with half of the instances considered ``easy'' and the other half considered ``hard''. The labelers labeled the ``easy'' images correctly with 100\% accuracy. The labelers labeled the ``hard'' images correctly according to whether they were ``good'' ($p_{correct} = .95$) or ``bad'' ($p_{correct}=.54$). The ratio of ``good'' to ``bad'' labelers is 25:1. The score is measured as the proportion of correctly estimated labels, reported as the error rate, $error = 1 - accuracy$. Whitehill et al. reported the average of 20 experiments; we report the mean of 50,000 experiments (Table \ref{whitehill2009-2}).

In the image difficulty modeling experiment, OpinionRank is able to recover the correct label in 100\% of cases. This is due to the parameters of the experiment. The ``easy'' images, being labeled with 100\% reliability, are heuristically irrelevant to the performance of the OpinionRank algorithm, as all voters will provide the same (correct) label. Therefore, regardless of the relative reliability eigenvector, the weighted sum will always be the correct label. The ``hard'' images, on the other hand, are also simple for OpinionRank, due to the labeling schema. With such a large majority of the labelers being ``good'', OpinionRank builds very strong recommendation relations between the ``good'' labelers, so the ``bad'' labelers are overruled when they are wrong (the 5\% of the time that the ``good'' labelers are wrong is also easily ignored). 

Whitehill et al.'s experiment was heavily biased toward ``good'' labelers (a reasonable scenario in the context of human annotators). We extended the experiment to smaller proportions of ``good'' to ``bad'' labelers; with 25, 40, and 50 (out of 50) labelers being ``bad'', we found error rates of 0\%, 1\%, and 14\%, respectively. These results suggest that OpinionRank is robust even against labeler pools with a high density of unreliable labelers.

\subsubsection{Stability Under Various Starting Points}
The third experiment simulates the labeling of $n = 2000$ instances by 20 labelers, with $\alpha_i \sim U[0, 4]$ and $\log(\beta_j)\sim U[0, 3]$. Under the assumptions of the authors' generative model, these parameters represent a large variance in the difficulty-expertise spectrum, and so the experiment tests algorithmic stability across a broad range of starting conditions. The scores are reported as the mean and standard deviation of label accuracy scores. Whitehill et al. reported the mean and standard deviation over 50 experiments; we report the mean and standard deviation over 50,000 experiments. 

\begin{table}
    \centering
    \caption{Mean error rate for estimated labels when modeling image difficulty (Section \ref{whitehill-ex-2}). Baseline results reproduced from Section 4 of Whitehill et al. \cite{whitehill}.}
    \begin{tabular}{cc}
        \toprule
        \textbf{Method} & \textbf{Error} \\ 
        \midrule
        Majority vote & 11.2\% \\ 
        Dawid \& Skene & 8.4\% \\ 
        GLAD & 4.5\% \\ 
        \textbf{OpinionRank} & \textbf{0.0\%} \\
        \bottomrule
    \end{tabular}
    \label{whitehill2009-2}
\end{table}




Similarly to the second experiment, OpinionRank achieves a perfect score on the authors' stability test (compared to mean of 85.84\% and standard deviation of 0.024\% for GLAD). Because the test draws the labeler expertise from $\alpha \sim U[0, 4]$, all labelers have expertise greater than random guessing. With a pool of 20 labelers, OpinionRank is able to consistently discover the \textit{best} labelers, even within this pool of above-average labelers, and  extract the correct labels. Notably, it achieves this performance without the computationally costly need to estimate the precise parameters of each labeler. OpinionRank demonstrates that only the \textit{relative} expertise is needed, as long as the Condorcet criterion is obeyed and the average expertise is greater than random chance \cite{condorcet-extended}.

\subsection{The Multidimensional Wisdom of Crowds}
We also implemented two experiments from Welinder et al. \cite{welinder}. The first experiment evaluates the OpinionRank algorithm on the authors' proposed label generation model (as described in Section \ref{welinder-setting}); the second experiment evaluates OpinionRank on a real-world dataset of human annotations. 

\subsubsection{Multidimensional Labeling Model}
\label{welinder-experiment}
\begin{figure}
    \centering
    \includegraphics[width=\linewidth]{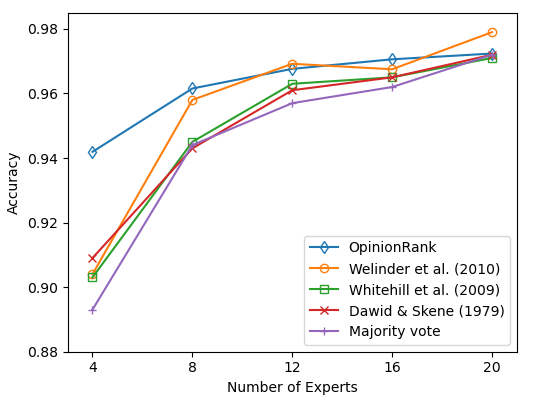}
    \caption{Performance of the OpinionRank algorithm in predicting $z_i \in [0, 1]$ on the data from Section \ref{welinder-experiment}, compared to majority voting, the EM model of Dawid and Skene \cite{dawid1979maximum}, the GLAD model of Whitehill et al. \cite{whitehill}, and the model of Welinder et al. \cite{welinder}. Baseline results reproduced from Figure 3(c) in Welinder et al. \cite{welinder}.}
    \label{welinder-1}
\end{figure}

We reproduce the conditions of Welinder et al.'s modeling experiment by generating data following the assumptions of the model (as described in Section \ref{welinder-setting}). Following the experimental setup in \cite{welinder}, we: 

\begin{itemize}
    \item set the number of data instances as $n=500$.
    \item assign $w_j = 1$ with probability $.99$ and $w_j = -1$ with probability .01, to simulate adversarial annotators.
    
    \item draw $\tau_j \sim \mathcal{N}(0, \sigma = 0.5)$.
    
    \item draw the noise parameter $\sigma_j \sim \text{Gamma}(1.5, 0.3)$.
    
    \item set the generative parameter $\theta_z = 0.5$. This value is  different from the  $\theta_z = 0.8$  as reported in \cite{welinder}. However, this change was done  after examining the code available at the authors' GitHub page\footnote{Welinder, P., ``Caltech UCSD Binary Annotation Model,'' Github, 2012. Available at https://github.com/welinder/cubam.} and discovering that they have in fact set this parameter to 0.5.
\end{itemize} 

Welinder et al. report the average over 40 experiments; we report the mean of 50,000 experiments (Figure \ref{welinder-1}). The annotation model of Welinder et al. is considerably more complex than that of Whitehill et al. Despite this complexity, OpinionRank achieves accuracy above 94\% across all experiments. We note that while all of the algorithms being compared eventually achieve accuracies $>96\%$, OpinionRank strongly outperforms the other algorithms at lower numbers of experts.




\subsubsection{Waterbirds Dataset}
\label{waterbirds}
\begin{table}
    \centering
    \caption{Percent correct labels on the Waterbirds dataset (Section \ref{waterbirds}). Performance of other algorithms reproduced from Section 5.2 of Welinder et al. \cite{welinder}.}
    \begin{tabular}{cc}
        \toprule
        \textbf{Method} & \textbf{Percent Correct} \\ 
        \midrule
        Majority voting & 68.3\% \\
        GLAD & 60.4\% \\ 
        Welinder et al. & 75.4\% \\
        \textbf{OpinionRank} & \textbf{86.7\%} \\ 
        \bottomrule
    \end{tabular}
    \label{welinder-2}
\end{table}
We evaluate OpinionRank on the real-world Waterbirds dataset constructed by the authors. Using Amazon Mechanical Turk, the authors asked 53 human annotators to provide labels on a set of 240 images. The images consisted of 50 photographs each of Mallards, American Black Ducks, Canadian Geese, and Red-necked Grebes, as well as 40 additional images featuring no birds. The annotators provided binary labels according to whether, in their opinion, each image contained a picture of a duck (only Mallards and American Black Ducks are positive classes). Of the 53 annotators, only 25 provided labels for all images; the other 28 annotators omitted between 40 and 200 labels. 

On this real-world dataset, OpinionRank predicts the correct label for 86.7\% of the images (Table \ref{welinder-2}). OpinionRank outpaces majority vote at 68.3\% accuracy, GLAD at 60.4\% accuracy, and the authors' own Bayesian generative model at 75.4\% accuracy.

\subsection{Combining Soft Decisions of Several Unreliable Experts}
\label{goldberger-ex}
\begin{figure}
    \centering
    \includegraphics[width=\linewidth]{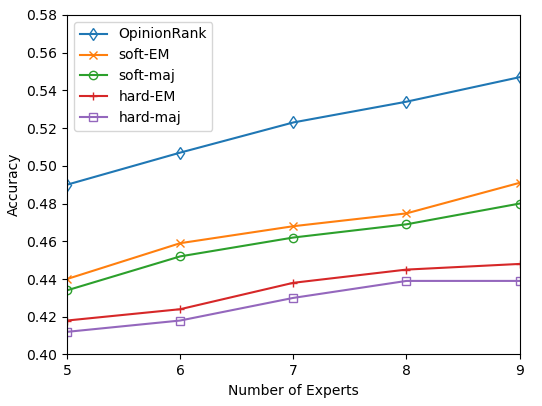}
    \caption{Accuracy of the OpinionRank algorithm as a function of the number of experts on the three-class soft opinions problem (Section\ref{goldberger-ex}) compared to the EM models of Goldberger. soft-EM, soft-maj, hard-EM, and hard-maj results reproduced from Figure 1 in Goldberger \cite{goldberger}.}
    \label{goldberger-1}
\end{figure}

We reproduced the soft-decision modeling experiment under the same conditions described in Goldberger \cite{goldberger}. Following Section \ref{goldberger-setting}, $n=200$ instances are generated and assigned random labels drawn from a pool of three classes. For each expert $j$, its reliability is sampled uniformly on the interval $[0.4, 0.7]$. After each expert's opinion $y_{ij}$ is modeled (as described in Section \ref{goldberger-setting}), opinions are obfuscated by first sampling a multinomial distribution $U_{ij}$ from the flat Dirichlet distribution, before transforming $y_{ij}$ into $U_{ij}$ following
\begin{displaymath}
    \hat{y} = \arg\max_{a\in A} U_{ij}[(y_{ij} + z - a) \mod |A|],\ \forall a\in A,
\end{displaymath}
where $z \in A$ is randomly sampled from $U_{ij}$. Because OpinionRank requires ``hard'' labels, we utilized Goldberger's hard-decision process, which takes the argmax of the soft label information over the set of classes, before providing the labels to the algorithm. 




Goldberger reports the mean of 100 experiments; we report the mean of 50,000 experiments. As seen in Figure \ref{goldberger-1}, OpinionRank outperforms the Goldberger's extended EM algorithm by a considerable margin, achieving at least 49\% accuracy (with only 5 experts), climbing monotonically up to 55\% accuracy (with 9 experts). 




\subsection{Empirical Runtime Analysis}
\label{runtime-ex}
We have also performed an empirical study of the wall clock runtime of the OpinionRank algorithm. We parameterized the experiment over $s$, the number of unreliable label sources (i.e., experts, labelers), and $n$, the total number of data instances. We vary $s$ between 1 and 100 sources, and $n$ between 10 and 1000 instances. All experiments were performed on a consumer-grade AMD Ryzen 3900X 3.8 GHz 12-core processor with 32 GB of memory. Each experiment occurred on a single processing thread. 

\begin{figure}
    \centering
    \includegraphics[width=\linewidth]{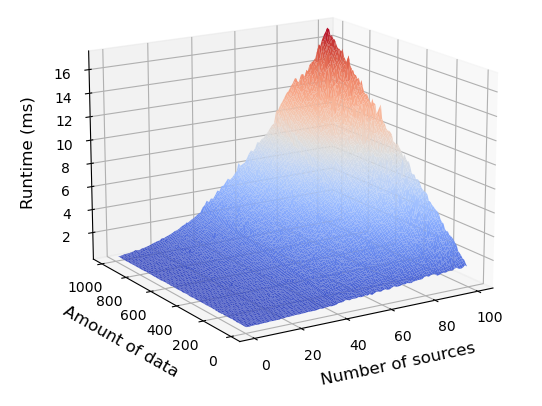}
    \caption{Wall clock runtime analysis of the OpinionRank algorithm.}
    \label{runtime}
\end{figure}

We generated an arbitrary $s \times n$ binary array of randomly generated class membership opinions. This array is passed to the OpinionRank algorithm, and we measure the time required for the algorithm to return its array of weighted class membership scores. We repeated this procedure 100 times for each set of parameters, whose average runtimes are depicted in Figure \ref{runtime}. We observe that the runtime of OpinionRank scales linearly with the amount of data, and quadratically with the number of sources. We note that the worst-case runtime, with $s=100$ and $n=1000$, is only 16.684 milliseconds. Scaling the amount of data up to 1 million instances only increased the average runtime to 17.712 seconds.

\section{Conclusions and Future Work}
Label-based supervised and semi-supervised learning requires trust in the veracity of the label information accompanying training datasets. Although crowdsourcing is an attractive and popular method for distributed gathering of such label information, it suffers from the inherent unreliability of individual annotators. We considered the problem of extracting reliable, ground truth labels from ensembles of label sources with unknown expertise. We provided a brief overview of the three conventional models of learning under unreliable labels, as well as the hand-crafted algorithmic solutions designed for each model. However, we noted that each of these solutions depends strongly on the correct parameterization of the label generation dynamics. To address this drawback, we proposed OpinionRank, a model-free, graph-based spectral method for tackling the problem of unreliable experts, which does not depend on any assumptions about the learning environment. OpinionRank is a very fast, scalable, efficient, and interpretable algorithm that can integrate label information from both human and machine label sources, while remaining robust to sources that may be highly unreliable or even adversarial. We evaluated OpinionRank on each of the settings considered, and found that it performed at least as well as---and often better than---the hand-crafted algorithms across all models of the labeling paradigm. OpinionRank performs remarkably well even across highly complex parameterizations of unreliable label generation, suggesting that complete knowledge of the label generation dynamics is not necessary for obtaining more accurate labels. Furthermore, OpinionRank requires considerably fewer computational resources than the expectation-maximization, Bayesian generative models, or deep neural network approaches proposed in the literature. OpinionRank can be seamlessly incorporated into any supervised or semi-supervised learner as a preprocessing step to increase the trust that the learner has for the labels on which it is trained.

To the best of our knowledge, our work is the first attempt to use Perron-Frobenius eigenvector ranking for evaluating relative reliability of unreliable experts. Because the method is well-researched in the context of the PageRank algorithm, we are interested in investigating even more powerful improvements upon this approach, such as the case where a prior is somehow known on a particular ground truth label (such as via querying an oracle). Other intriguing avenues of research include incorporating OpinionRank into more well-known classifiers, its applications in the context of other ensemble learning systems, and its potential as a tool for detecting and relabeling adversarial examples. 

\section*{Acknowledgements}
This work was supported by the U.S. Department of Education, GAANN grant no. P200A180055.

\bibliographystyle{ieeetr}
\bibliography{references.bib}

\begin{thebibliography}{10}

\bibitem{imagenet}
J.~{Deng}, W.~{Dong}, R.~{Socher}, L.-J. {Li}, {Kai Li}, and {Li Fei-Fei},
  ``Imagenet: A large-scale hierarchical image database,'' in {\em 2009 IEEE
  Conf. on Computer Vision and Pattern Recognition}, pp.~248--255, 2009.

\bibitem{snow2008cheap}
R.~Snow, B.~O'Connor, D.~Jurafsky, and A.~Y. Ng, ``Cheap and fast---but is it
  good? evaluating non-expert annotations for natural language tasks,'' in {\em
  Proc. of the Conf. on Empirical Methods in Natural Language Processing},
  EMNLP '08, p.~254–263, 2008.

\bibitem{wang2013}
A.~Wang, C.~D. Hoang, and M.-Y. Kan, ``Perspectives on crowdsourcing
  annotations for natural language processing,'' {\em Language Resources and
  Evaluation}, vol.~47, p.~9–31, Mar. 2013.

\bibitem{guo2019}
B.~Guo, H.~Chen, Y.~Liu, C.~Chen, Q.~Han, and Z.~Yu, ``From crowdsourcing to
  crowdmining: Using implicit human intelligence for better understanding of
  crowdsourced data,'' {\em World Wide Web}, vol.~23, no.~2, pp.~1101--1125,
  2020.

\bibitem{sheng2008}
V.~S. Sheng, F.~Provost, and P.~G. Ipeirotis, ``Get another label? improving
  data quality and data mining using multiple, noisy labelers,'' in {\em Proc.
  of the 14th ACM SIGKDD Int. Conf. on Knowledge Discovery and Data Mining},
  KDD '08, p.~614–622, 2008.

\bibitem{xintong2014}
G.~Xintong, W.~Hongzhi, Y.~Song, and G.~Hong, ``Review: Brief survey of
  crowdsourcing for data mining,'' {\em Expert Syst. Appl.}, vol.~41,
  p.~7987–7994, Dec. 2014.

\bibitem{imagenet-done}
L.~Beyer, O.~J. H{\'e}naff, A.~Kolesnikov, X.~Zhai, and A.~v.~d. Oord, ``Are we
  done with imagenet?,'' {\em arXiv preprint arXiv:2006.07159}, 2020.

\bibitem{imagenet-relabeling}
S.~Yun, S.~J. Oh, B.~Heo, D.~Han, J.~Choe, and S.~Chun, ``Re-labeling imagenet:
  from single to multi-labels, from global to localized labels,'' {\em arXiv
  preprint arXiv:2101.05022}, 2021.

\bibitem{condorcet}
M.~J. A.~N. de~Caritat Marquis~de Condorcet, {\em Essai sur l'application de
  l'analyse {\`a} la probabilit{\'e} des d{\'e}cisions rendues {\`a} la
  pluralit{\'e} des voix}.
\newblock L'imprimerie royale, 1785.

\bibitem{democratic}
R.~G. Kazmann, ``Democratic organization: A preliminary mathematical model,''
  {\em Public Choice}, vol.~16, no.~1, pp.~17--26, 1973.

\bibitem{condorcet-unreliable}
B.~Grofman, ``A comment on ‘democratic theory: A preliminary mathematical
  model.’,'' {\em Public Choice}, vol.~21, no.~1, pp.~99--103, 1975.

\bibitem{condorcet-extended}
G.~Owen, B.~Grofman, and S.~L. Feld, ``Proving a distribution-free
  generalization of the condorcet jury theorem,'' {\em Math. Soc. Sci.},
  vol.~17, no.~1, pp.~1--16, 1989.

\bibitem{condorcet-multiclass}
C.~List and R.~E. Goodin, ``Epistemic democracy: Generalizing the condorcet
  jury theorem,'' {\em J. Political Philos.}, vol.~9, no.~3, pp.~277--306,
  2001.

\bibitem{dawid1979maximum}
A.~P. Dawid and A.~M. Skene, ``Maximum likelihood estimation of observer
  error-rates using the em algorithm,'' {\em J. R. Stat. Soc. Ser. C (Appl.
  Stat.)}, vol.~28, no.~1, pp.~20--28, 1979.

\bibitem{venus}
P.~Smyth, U.~Fayyad, M.~Burl, P.~Perona, and P.~Baldi, ``Inferring ground truth
  from subjective labelling of venus images,'' in {\em Adv. in NeurIPS},
  vol.~7, pp.~1085--1092, 1995.

\bibitem{whitehill}
J.~Whitehill, P.~Ruvolo, J.~Bergsma, T.~Wu, and J.~Movellan, ``Whose vote
  should count more: Optimal integration of labels from labelers of unknown
  expertise,'' in {\em Adv. in NeurIPS}, 2009.

\bibitem{ensemble-linear}
A.~{Alush} and J.~{Goldberger}, ``Ensemble segmentation using efficient integer
  linear programming,'' {\em IEEE Trans. Pattern Anal. Mach. Intell.}, vol.~34,
  no.~10, pp.~1966--1977, 2012.

\bibitem{zhou2014aggregating}
D.~Zhou, Q.~Liu, J.~C. Platt, and C.~Meek, ``Aggregating ordinal labels from
  crowds by minimax conditional entropy,'' in {\em Proc. of the 31st Int. Conf.
  on Machine Learning - Volume 32}, ICML'14, p.~II–262–II–270, 2014.

\bibitem{gao2013minimax}
C.~Gao and D.~Zhou, ``Minimax optimal convergence rates for estimating ground
  truth from crowdsourced labels,'' {\em arXiv preprint arXiv:1310.5764}, 2013.

\bibitem{rodrigues2013}
F.~Rodrigues, F.~Pereira, and B.~Ribeiro, ``Learning from multiple annotators:
  Distinguishing good from random labelers,'' {\em Pattern Recognit. Lett.},
  vol.~34, p.~1428–1436, Sept. 2013.

\bibitem{yan2014learning}
Y.~Yan, R.~Rosales, G.~Fung, R.~Subramanian, and J.~Dy, ``Learning from
  multiple annotators with varying expertise,'' {\em Mach. Learn.}, vol.~95,
  no.~3, pp.~291--327, 2014.

\bibitem{zhang2018}
J.~Zhang and X.~Wu, ``Multi-label inference for crowdsourcing,'' in {\em Proc.
  of the 24th ACM SIGKDD Int. Conf. on Knowledge Discovery \& Data Mining},
  p.~2738–2747, 2018.

\bibitem{ghosh}
A.~Ghosh, S.~Kale, and P.~McAfee, ``Who moderates the moderators? crowdsourcing
  abuse detection in user-generated content,'' in {\em Proc. of the 12th ACM
  SIGecom}, EC '11, p.~167–176, 2011.

\bibitem{dalvi}
N.~Dalvi, A.~Dasgupta, R.~Kumar, and V.~Rastogi, ``Aggregating crowdsourced
  binary ratings,'' in {\em Proc. of the 22nd Int. Conf. on World Wide Web},
  WWW '13, p.~285–294, 2013.

\bibitem{chen}
X.~Chen, Q.~Lin, and D.~Zhou, ``Optimistic knowledge gradient policy for
  optimal budget allocation in crowdsourcing,'' in {\em Proc. of the 30th Int.
  Conf. on Machine Learning}, no.~3 in Proceedings of Machine Learning
  Research, pp.~64--72, PMLR, 17--19 Jun 2013.

\bibitem{liu}
Q.~Liu, J.~Peng, and A.~T. Ihler, ``Variational inference for crowdsourcing,''
  in {\em Adv. in NeurIPS}, vol.~25, pp.~692--700, 2012.

\bibitem{raykar}
V.~C. Raykar, S.~Yu, L.~H. Zhao, G.~H. Valadez, C.~Florin, L.~Bogoni, and
  L.~Moy, ``Learning from crowds,'' {\em J. Mach. Learn. Res.}, vol.~11,
  no.~43, pp.~1297--1322, 2010.

\bibitem{welinder}
P.~Welinder, S.~Branson, P.~Perona, and S.~Belongie, ``The multidimensional
  wisdom of crowds,'' in {\em Adv. in NeurIPS}, vol.~23, pp.~2424--2432, 2010.

\bibitem{zou}
J.~Y. Zou, D.~J. Hsu, D.~C. Parkes, and R.~P. Adams, ``Contrastive learning
  using spectral methods,'' in {\em Adv. in NeurIPS}, vol.~26, pp.~2238--2246,
  2013.

\bibitem{zhang}
Y.~Zhang, X.~Chen, D.~Zhou, and M.~I. Jordan, ``Spectral methods meet em: A
  provably optimal algorithm for crowdsourcing,'' {\em J. Mach. Learn. Res.},
  vol.~17, no.~102, pp.~1--44, 2016.

\bibitem{goldberger}
J.~{Goldberger}, ``Combining soft decisions of several unreliable experts,'' in
  {\em 2016 IEEE Int. Conf. on Acoustics, Speech and Signal Processing},
  pp.~2334--2338, 2016.

\bibitem{yin2017}
L.~Yin, J.~Han, W.~Zhang, and Y.~Yu, ``Aggregating crowd wisdoms with
  label-aware autoencoders,'' in {\em Proc. of the 26th Int. Joint Conf. on
  Artificial Intelligence}, pp.~1325--1331, 2017.

\bibitem{atarashi2018semi}
K.~Atarashi, S.~Oyama, and M.~Kurihara, ``Semi-supervised learning from crowds
  using deep generative models,'' in {\em Proc. of the AAAI Conf. on Artificial
  Intelligence}, vol.~32, 2018.

\bibitem{shi2020}
W.~Shi, V.~S. Sheng, X.~Li, and B.~Gu, {\em Semi-Supervised Multi-Label
  Learning from Crowds via Deep Sequential Generative Model}, p.~1141–1149.
\newblock 2020.

\bibitem{pagerank}
L.~Page, S.~Brin, R.~Motwani, and T.~Winograd, ``The pagerank citation ranking:
  Bringing order to the web.,'' Technical Report 1999-66, Stanford InfoLab,
  November 1999.

\bibitem{mihalcea2004textrank}
R.~Mihalcea and P.~Tarau, ``{T}ext{R}ank: Bringing order into text,'' in {\em
  Proc. of the 2004 Conf. on Empirical Methods in Natural Language Processing},
  pp.~404--411, July 2004.

\bibitem{erkan2004lexrank}
G.~Erkan and D.~R. Radev, ``Lexrank: Graph-based lexical centrality as salience
  in text summarization,'' {\em J. Artif. Intell. Res.}, vol.~22, p.~457–479,
  Dec. 2004.

\bibitem{burges2005learning}
C.~Burges, T.~Shaked, E.~Renshaw, A.~Lazier, M.~Deeds, N.~Hamilton, and
  G.~Hullender, ``Learning to rank using gradient descent,'' in {\em Proc. of
  the 22nd Int. Conf. on Machine Learning}, ICML '05, p.~89–96, 2005.

\bibitem{chakrabarti2008structured}
S.~Chakrabarti, R.~Khanna, U.~Sawant, and C.~Bhattacharyya, ``Structured
  learning for non-smooth ranking losses,'' in {\em Proc. of the 14th ACM
  SIGKDD Int. Conf. on Knowledge Discovery and Data Mining}, KDD '08,
  p.~88–96, 2008.

\bibitem{yue2007support}
Y.~Yue, T.~Finley, F.~Radlinski, and T.~Joachims, ``A support vector method for
  optimizing average precision,'' in {\em Proc. of the 30th ACM SIGIR
  Conference on Research and Development in Information Retrieval}, SIGIR '07,
  p.~271–278, 2007.

\bibitem{frobenius1912matrizen}
G.~Frobenius, {\em {\"U}ber Matrizen aus nicht negativen Elementen}.
\newblock K{\"o}nigliche Akademie der Wissenschaften Sitzungsber, K{\"o}n,
  1912.

\bibitem{seeley1949net}
J.~R. Seeley, ``The net of reciprocal influence. a problem in treating
  sociometric data,'' {\em Can. J. Exp. Psychol.}, vol.~3, p.~234, 1949.

\bibitem{Keener1993ThePT}
J.~Keener, ``The perron-frobenius theorem and the ranking of football teams,''
  {\em SIAM Review}, vol.~35, pp.~80--93, 1993.

\bibitem{chen2019spectral}
Y.~Chen, J.~Fan, C.~Ma, and K.~Wang, ``Spectral method and regularized mle are
  both optimal for top-k ranking,'' {\em Ann. Stat.}, vol.~47, no.~4, p.~2204,
  2019.

\end{thebibliography}

\end{document}